# MEAL-TAKING ACTIVITY MONITORING IN THE ELDERLY BASED ON SENSOR DATA: COMPARISON OF UNSUPERVISED CLASSIFICATION METHODS

Abderrahim Derouiche[1], Damien Brulin[1], Eric Campo[1] and Antoine Piau[2]

[1]LAAS-CNRS, University of Toulouse
CNRS, UPS, UT2J, Toulouse, France

[2]Center for Epidemiology and Population Health Research
CHU Toulouse, Toulouse, France

**ABSTRACT**

In an era marked by a demographic change towards an older population, there is an urgent need to improve nutritional monitoring in view of the increase in frailty. This research aims to enhance the identification of meal-taking activities by combining K-Means, GMM, and DBSCAN techniques. Using the Davies-Bouldin Index (DBI) for the optimal meal-taking activity clustering, the results show that K-Means seems to be the best solution, thanks to its unrivalled efficiency in data demarcation, compared with the capabilities of GMM and DBSCAN. Although capable of identifying complex patterns and outliers, the latter methods are limited by their operational complexities and dependence on precise parameter configurations. In this paper, we have processed data from 4 houses equipped with sensors. The findings indicate that applying the K-Means method results in high performance, evidenced by a particularly low Davies-Bouldin Index (DBI), illustrating optimal cluster separation and cohesion. Calculating the average duration of each activity using the GMM algorithm allows distinguishing various categories of meal-taking activities. Alternatively, this can correspond to different times of the day fitting to each meal-taking activity. Using K-Means, GMM, and DBSCAN clustering algorithms, the study demonstrates an effective strategy for thoroughly understanding the data. This approach facilitates the comparison and selection of the most suitable method for optimal meal-taking activity clustering.

## 1. INTRODUCTION

With the increase of elderly population in the world's comes the imperative to improve their life quality (Shlisky,2017), paying particular attention to the management of their daily nutritional needs (Holman,2020). Meticulous monitoring of the meal-taking habits of the elderly is essential to meet their nutritional needs and avoid the dangers associated with a poor nutrition. These dangers include sarcopenia, which compromises mobility; nutritional fragility, which increases susceptibility to infection and delays healing; and the aggravation of age-related chronic diseases such as cardiovascular disease, type 2 diabetes and certain forms of cancer (Volkert,2022). Digital technology is emerging as an effective method of monitoring meal-taking habits in real time (Suryadevara,2023). Aging is accompanied by a number of physiological changes that can affect nutritional status, such as changes in metabolism, reduced appetite, altered taste perception and difficulties in consuming food (Ghias,2023).

The use of technological innovations to monitor diet is a new approach to detect and predict risks (Burrows,2019). Thanks to technological advances, it is possible to observe eating habits as they happen, to identify nutritional problems quickly and to remedy them (Sempionatto,2021). Devices such as apps and smart objects play a key role in monitoring food and fluid intake, making it easier to adapt diets to the unique nutritional needs of the elderly. In this way, they ensure that their meal-taking habits correspond more closely to their nutritional needs (Armand, 2024).The use of digital technology in food management for the elderly offers a pioneering solution to the challenges posed by an aging population (Sakib,2023). This method not only raises the standard of living of the elderly through good nutrition, but also minimizes the risks of illness linked to poor eating habits and promotes a culture of healthy aging (Piccoli,2023). The integration of technological solutions in the nutritional care of the elderly marks a major step forward to a comprehensive and effective management of their nutritional health in the face of an aging society (Khalifa,2024).

In this context, the CART France initiative project (Derouiche, 2023) is emerging as a leader in the application of artificial intelligence to refine the identification of meal-taking habits in residential settings. Based on data collected as part of the CART project, this paper presents an innovative strategy that integrates K-Means (Arora,2016), Gaussian mixture model (GMM) (Cassimon,2023), and density-based spatial clustering of applications with noise (DBSCAN) (Ahmed,2016) algorithms for analyzing behavioral data related to residents' meal-taking habits. The Davies-Bouldin Index (DBI) was used to determine the optimal number of clusters to form (Made,2022). Once this has been done, the DBI is used again to determine the best clustering value and validate it by comparing the three clustering methods.

## 2. RELATED WORKS

In this section, we highlight three recent studies that use emerging technologies such as machine learning, deep learning, and wearable sensors to improve nutritional prediction, physical activity monitoring, and meal-taking identification. These research efforts demonstrate significant progress in accurately predicting nutritional intake from food images, improving activity and meal-taking recognition via wearable sensors, and understanding how older people adapt to maintain their nutrition despite age-related challenges.

B. Wang et al (Wang,2023) present a novel method for nutrition prediction from food images, which overcomes the challenges of direct regression by using a coarse-to-fine framework and structural linear smoothing loss. This approach significantly improves prediction accuracy over several benchmarks, illustrating its effectiveness in handling the wide range of nutrient values in foods and the complexity of food image data. The study highlights the advances of the method while recognizing the need for further research to address the variability of food intake environments and the potential problems associated with food waste management due to excessive control of nutrient intake.

G. Hussain et al (Hussain,2019) present a study on a wearable sensor system combining a smartwatch and necklace with a piezoelectric sensor for automated activity and food recognition in preventive healthcare. It achieves high accuracy in recognizing eight physical activities (e.g. going downstairs, eating, going upstairs, walking…) and six food categories, demonstrating the system's effectiveness. However, the study recognizes challenges such as the variability of food characteristics and the need for further improvement in sensor technology and algorithm optimization to increase recognition accuracy and adaptability in various real-life settings.

E. Setiawati et al. (Setiawati,2024) carried out a study using K-Means, K-Medoid, and DBSCAN algorithms to cluster obesity data and uncover patterns and relationships. The evaluation, based on the Davies-Bouldin Index (DBI), revealed that the K-Means algorithm with K2 clusters performed the best, achieving a DBI value of 0.604. The K-Medoid algorithm also performed well with a DBI value of 0.614, while the DBSCAN algorithm showed optimal performance with K3 clusters, achieving a DBI value of 1.040. The study concludes that the K-Means algorithm is the most effective for clustering obesity data. Nevertheless, limitations of the study include the exclusive use of the Davies-Bouldin Index for validation and the potential variability of results with different datasets.

These few illustrative examples show different approaches for accessing nutritional prediction and physical activity monitoring in the elderly. In the next section, we present the methodology used in our study.

## 3. MATERIAL AND METHOD

### 3.1 Data Collecting

In this study, the initial phase consists of collecting raw data from passive infrared motion sensors and contact sensors installed in the homes of elderly people living alone (average age of 82 with no particular pathology). These sensors generate binary data when activated, with the data collected over a 12-month period. The compiled database is in CSV format, and constitutes a large dataset. The next step is to develop a new algorithm designed to identify and monitor various types of nutritional activities within the home, focusing particularly on data from the kitchen or dining room, as shown in Figure 1. Data clustering is performed using three algorithms: K-Means, GMM, and DBSCAN.

Figures 2 and 3 show the process flow charts. The first flowchart details the steps involved in identifying and recording food-related activities from a CSV file, exploring the data to detect specific actions and calculating the duration of these activities. The second flowchart describes a general process for classifying data related to food activity using three different clustering algorithms (K-Means, GMM, and DBSCAN) to assess the accuracy of the results.

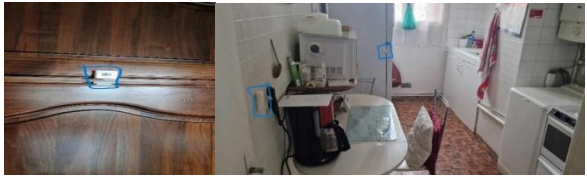

FIG.1. Contact and Motion Sensors in a Kitchen.

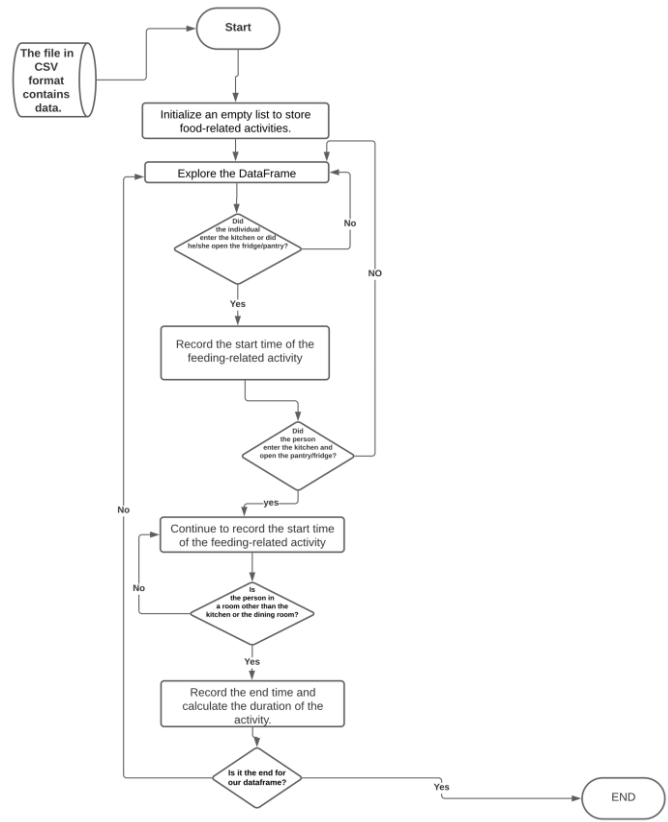

FIG.2. Nutrition Activity Tracking Flowchart.

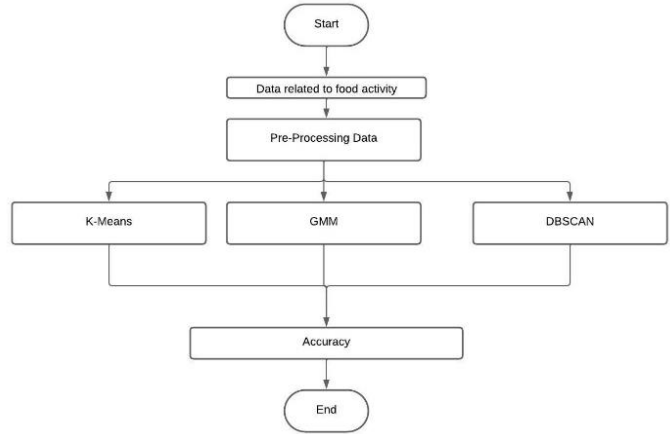

FIG.3. Process flow for Nutrition Activity Clustering Analysis.

## 3.2 Clustering

Unsupervised clustering (Sabita,2021) appears to be an ideal method for discovering the intrinsic structures of food data, without the influence of predefined labels. It reveals unexpected classifications, such as the distinction between different types of food activities (meals and snacks), while also revealing hidden patterns. Furthermore, these methods objectively determine the optimal number of clusters based on mathematical criteria, thereby adjusting the complexity of the model to the relevance of the data. In our particular context, unsupervised clustering is proving very effective in establishing thresholds based on observed durations, making it easier to distinguish between meals and snacks.

### 3.2.1 K-Means

K-Means Clustering is a method used in data analysis and mining that operates in an unsupervised manner. It adopts a partitioning strategy to cluster data, aiming to group similar features within clusters while segregating dissimilar ones into separate clusters (Yuan,2019).

The main objective of K-Means is to enhance similarity within clusters while reducing it between distinct clusters. The crucial factor used to amplify data similarity within a cluster is the distance function, which evaluates the similarity of data by measuring the closest distance to the cluster center, as illustrated in equation 1.

$$D_e = \sqrt{(x_i - s_i)^2 (y_i - t_i)^2} \quad (1)$$

### 3.2.2 GMM

Gaussian Mixture Model, commonly known as GMM, is a statistical framework represented by a mixture of densities. It is generally used to parametrically estimate the distribution of random variables, by conceptualizing them as a combination of several Gaussian components, also known as kernels. The aim is to determine the parameters of each Gaussian component, such as the mean, variance, and amplitude, optimizing them based on the maximum likelihood principle in order to obtain a distribution close to the desired one (El Attar,2012).

Gaussian Mixture Models (GMM) are limited by their sensitivity to initial conditions, the difficulty of choosing the correct number of Gaussian components, and the restrictive assumption that the data follow normal distributions (as illustrated in equation 2), which may not be valid for all data.

$$g(x, \Phi) = \sum_{k=1}^{g} \pi_k F(x, \theta_k) \quad (2)$$

$g(x, \Phi)$: This is the overall GMM probability density function for a data point x, given the model parameters.
$g$: The number of components (Gaussian distributions) in the mixture.
$\sum_{k=1}^{g}$: This is the summation sign, which indicates that the next term must be added to all the g components of the mixture.

$\pi_k$: The mixing coefficient for the Kth component. It represents the weight of the Kth Gaussian in the mixture and satisfies $0 \leq \pi_k \leq 1$ and $\sum_{k=1}^{g} \pi_k = 1$. This ensures that the total probability distribution sums to 1.

$F(x, \theta_k)$: The probability density function of the kth Gaussian component. This function is parameterized by $\theta_k$, which typically includes the mean and covariance of the kth Gaussian.

### 3.2.3 DBSCAN

DBSCAN stands out in scenarios where K-means and GMM show limitations, thanks to its ability to detect clusters of various shapes and skillfully manage outliers and noise. Unlike the need to predefine cluster numbers as in K-means and GMM, DBSCAN uses the density of the data to discover this parameter autonomously, making it uniquely suitable with datasets with complex and non-linear structures. Additionally, its ability to handle fluctuations in data density gives it a significant advantage in processing effectively with data of varying densities, a challenge often faced by conventional methods focused on identifying clusters of a specific geometric shape.

$$N_{Eps} = \{q \in D / dist(p,q) < Eps\} \quad (3)$$

$N_{Eps}$: This represents the Epsilon-neighborhood of a point p. It is the set of the points that fall within a radius Epsilon from point p.
q: A point within the dataset D.
D: The complete dataset in which clustering is being performed.
dist(p,q): The distance function between points p and q.
Eps: The Epsilon parameter, which is the radius of the neighborhood around point p.

## 4. RESULTS AND ANALYSIS

## 4.1. Comparison of clustering Algorithms

The Davies-Bouldin scores, computed using the K-Means approach as shown in Figure 4, reveal optimal quantities of clusters that vary from household to household, highlighting the diversity of behavior patterns related to food consumption. Household 2385 shows optimal clustering at 10 clusters, while household 2445 is optimal with 8 clusters. For household 2373, the data clearly give an optimal clustering with 10 clusters. For household 2356, the indicators suggest maximum clustering performance at 6 clusters as well.

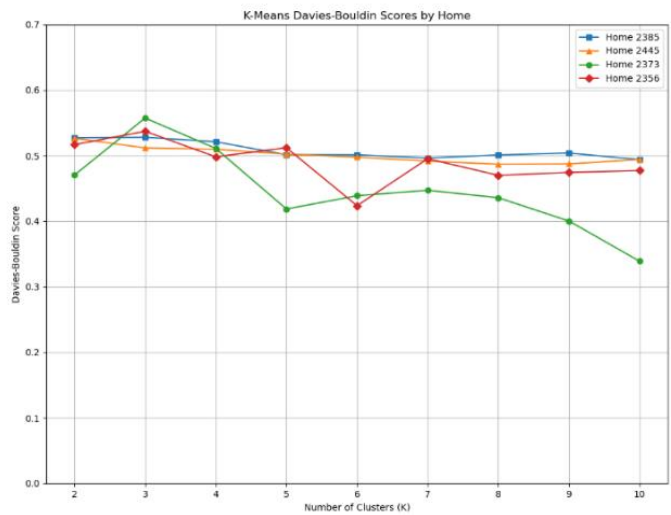

FIG. 4. K-Means Davies Bouldin scores for 4 houses.

The GMM clustering analysis and Davies-Bouldin scores presented in Figure 5 suggest that increasing the number of clusters often leads to more pronounced separations, particularly noticeable at K=10 for households 2385 and 2373, where the most distinct demarcations occur. Household 2445 finds its optimal clustering definition for K=9, while household 2356 is optimal with K=7.

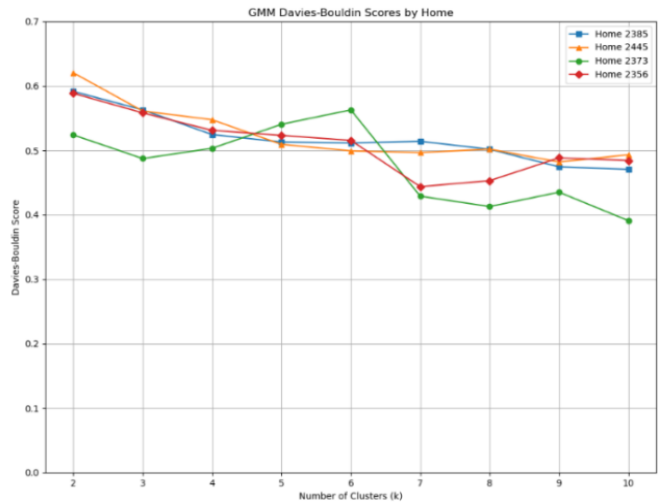

FIG. 5. GMM Davies Bouldin scores for 4 houses.

Davies-Bouldin scores derived from DBSCAN clustering, as shown in Figure 6, give optimal Eps values for discerning food-related activity clusters within households. Households 2385 and 2356 show better cluster separation for Eps values of 10 and 9 respectively. Household 2445 show better cluster separation for Eps value 4 and household 2373 show better cluster separation for EPS value 3.

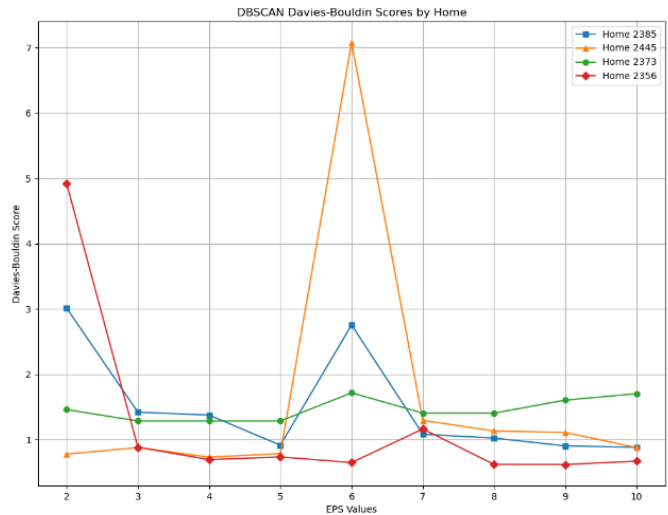

FIG. 6. DBSCAN Davies Bouldin scores for 4 houses.

These results indicate the diversity of meal-taking activity patterns and help determine the ideal number of activity types for each household.

## 4.2. Discussion

When comparing clustering methods to identify meal-taking activities, GMM and K-Means perform better as the number of clusters increases. K-Means, in particular, stands out for lower DBI scores when the number of clusters is higher, suggesting better differentiation between activities such as meals and snacks. The effectiveness of DBSCAN varies considerably with Eps values, requiring fine-tuning of parameters for optimal clustering. This comparative study shows that K-Means appears to be the most consistent method for distinguishing the types of meal-taking activities in different households (Figure 4). The complex and sometimes multimodal nature of meal-taking means that there is a wide range of different ways of eating. Leveraging the K-means algorithm to ascertain the ideal number of clusters, the GMM algorithm is subsequently used. This algorithm adeptly harnesses a collection of Gaussian curves, offering the agility needed to encompass the assorted variability. This methodology not only facilitates a detailed depiction of the multifaceted nature of meal-taking behaviors but also enables the computation of the mean duration associated with each habit-related activity. This approach allows us to estimate the means and variances of the Gaussians as shown in Figure 7. Calculating the mean duration of individual activities through the GMM algorithm facilitates the differentiation of diverse categories of meal-taking behaviors (4 categories of meal-taking behaviors shown in Figure 7). Alternatively, this could pertain to distinct times of the day associated with each meal-taking activity.

Although the analysis can identify clusters and trends, the practical application of these results to improve elderly nutrition requires additional interventions and monitoring steps beyond data analysis.

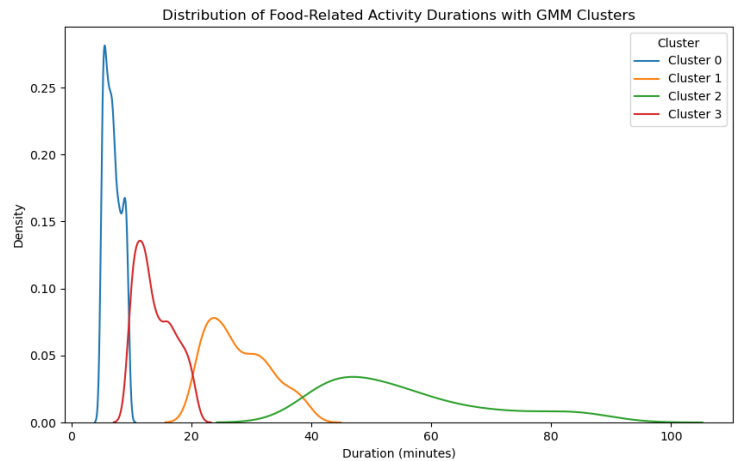

FIG. 7. Distribution of Nutrition-Related Activity Durations with GMM.

## 5. CONCLUSION

This research focuses on determining the optimal number of clusters in the context of monitoring meal-taking activity. Employing three different clustering algorithms (K-Means, GMM, DBSCAN) is an effective strategy that enables a thorough understanding of the data and facilitates the comparison and selection of the most suitable method. It concludes that the K-Means algorithm seems to be the best choice for clustering due to clear data separation and consistent performance across different households. GMM's flexibility is advantageous for complex eating patterns, while DBSCAN's outlier detection is crucial for identifying atypical meal-taking activities. However, K-Means may falter with complex cluster shapes and outliers; GMM faces high computational resources; and DBSCAN's performance is highly dependent on precise parameter tuning. Future alert systems for meal-taking activity changes could benefit from an overall approach combining these methods, complemented by time series and predictive analysis for comprehensive, forward-looking meal-taking activity monitoring.